\documentclass[conference]{IEEEtran}
\usepackage[T1]{fontenc}
\usepackage[utf8]{inputenc}
\usepackage{cite}
\usepackage{graphicx}
\usepackage{booktabs}
\usepackage{tabularx}
\usepackage{array}
\usepackage{url}
\usepackage{amsmath}
\usepackage{balance}
\usepackage[hidelinks]{hyperref}
\begin{document}

\title{Optimal Stop-Loss and Take-Profit Parameterization for Autonomous Trading Agent Swarm}
\author{
\IEEEauthorblockN{Nathan Li}
\IEEEauthorblockA{University of Oxford}
\and
\IEEEauthorblockN{Aikins Laryea}
\IEEEauthorblockA{Vela Research}
\and
\IEEEauthorblockN{Yigit Ihlamur}
\IEEEauthorblockA{Vela Research}
}
\maketitle

\begin{abstract}
Autonomous crypto trading systems often spend most of their design effort on finding entries, while exits are left to fixed rules that are rarely tested in a systematic way. This paper examines whether better stop-loss and take-profit settings can improve the performance of an autonomous trading agent swarm. Using more than 900 historical trades, we replay each trade under many alternative exit policies and compare results against the existing production setup. The study finds that exit design matters meaningfully: stronger configurations improve risk-adjusted performance and generally favor tighter loss limits, earlier profit capture, and closer trailing protection. The paper also discusses a key evaluation challenge: a purely chronological split was initially used, but the newest trades fell into an unusual war-driven market period that sharply distorted test results. To reduce the influence of that single episode, the main comparison was run on randomized data, with the drawbacks of doing so acknowledged explicitly. Overall, the paper presents a practical framework for tuning exit logic in a more disciplined and transparent way.
\end{abstract}


\section{Introduction}
Exit logic is often the least justified component of autonomous trading systems. Entry models are usually tuned and monitored closely, while stop-loss, trailing-stop, and take-profit rules are inherited from prior deployments or chosen heuristically. In high-volatility crypto markets, that practice can degrade realized performance through delayed loss cutting, premature profit erosion, or both.

The trading system studied here is not a single monolithic policy but an autonomous agent swarm. Multiple specialized agents monitor market state, generate candidate entries, and hand approved positions to a shared execution and risk layer. Exit calibration therefore matters at the portfolio level as well as the individual-trade level, because the same stop-loss and take-profit policy must operate consistently across heterogeneous agent behaviors.

This paper asks a narrow question: \emph{which exit-parameter combinations produce the best risk-adjusted performance on the available historical trade set?} The contribution is not a claim of global optimality. Instead, it is an empirical calibration exercise on a discrete search space, designed to identify better settings than the current defaults and to clarify which parts of the exit stack matter most.

The study contributes three items. First, it presents a reproducible counterfactual simulation framework for replaying historical trades under alternative exit rules. Second, it reports a full-grid comparison of 8,960 configurations, plus a second-stage refinement using ATR-based overlays and circuit-breaker logic. Third, it translates the results into a compact production recommendation while stating the limits of the evidence clearly.

Prior work on trailing stops shows that exit design can materially change return distributions, but that performance claims are highly sensitive to market regime and evaluation protocol \cite{glynn1995,dai2021,kaminski2014}. That makes this problem practically important but methodologically easy to overstate. The present paper therefore prioritizes clear reporting over aggressive inference.

\section{Autonomous Trading Agent Swarm}
The production system uses an agent-swarm architecture rather than a single trading model. A group of approximately 10 to 20 agents, each associated with a different model family or signal perspective, evaluates the market on a recurring schedule. In broad terms, the swarm refreshes roughly every 15 minutes by ingesting a top-trader signal and converting it into candidate trade opportunities.

The agents act as a distributed idea-generation layer rather than as fully independent execution systems. Candidate positions produced by the swarm are filtered through a shared execution and risk layer that enforces portfolio-level constraints and applies a common exit policy. This high-level description is intentionally brief: it is sufficient for understanding why exit calibration matters here, without disclosing proprietary details of the entry logic or coordination mechanism.

That shared framework is exactly why exit calibration is important in this paper. Once a position is opened, the exit logic does not distinguish which agent originated the trade. Stop-loss, trailing-stop, partial-take-profit, ATR overlay, and stale-close rules are therefore applied as a unifying policy across the swarm. The calibration problem is to identify exit parameters that work well across the aggregate distribution of swarm-generated trades rather than overfitting to a single entry style.

\section{Data and Method}
The dataset contains over 900 closed trades for which complete price-path snapshots were available. The baseline production configuration in the first-pass search is a 25\% stop-loss, 3\% trailing activation, 2\% trailing distance, 5\% partial take-profit threshold, 50\% partial take-profit fraction, and 24-hour stale close. For each trade, the simulator replays the observed price path and applies a candidate set of exit rules in chronological order. The resulting realized return is then aggregated across trades.

The first-pass grid varies six parameters: stop-loss, trailing activation, trailing distance, partial take-profit threshold, partial take-profit fraction, and stale-position duration. Table~\ref{tab:grid} summarizes the search ranges. This stage produces 8,960 candidate configurations. The five strongest first-pass configurations are then refined in a second pass using ATR-based stop and take-profit multipliers and a circuit-breaker overlay.

\begin{table}[t]
\caption{Parameter ranges used in the two-stage search.}
\label{tab:grid}
\centering
\scriptsize
\setlength{\tabcolsep}{3pt}
\renewcommand{\arraystretch}{1.08}
\begin{tabularx}{0.9\linewidth}{@{}>{\raggedright\arraybackslash}p{0.40\linewidth}>{\raggedright\arraybackslash}X@{}}
\toprule
Parameter & Values tested \\
\midrule
Stop-loss (\%) & 5, 10, 15, 20, 25, 30, 50 \\
Trailing activation (\%) & 3, 5, 8, 10, 15 \\
Trailing distance (\%) & 2, 3, 5, 8 \\
Partial take-profit threshold (\%) & 5, 10, 15, 20 \\
Partial take-profit fraction & 25, 33, 50, 75 \\
Stale close & 12 h, 24 h, 48 h, 72 h \\
ATR stop multiplier & 1.0$\times$, 1.5$\times$, 2.0$\times$, 2.5$\times$, 3.0$\times$ \\
ATR take-profit multiplier & 2.0$\times$, 3.0$\times$, 4.0$\times$, 6.0$\times$ \\
Circuit-breaker loss threshold & 2, 3, 4, 5 consecutive losses \\
Circuit-breaker reduction factor & 0.25$\times$, 0.50$\times$, 0.75$\times$, disabled \\
\bottomrule
\end{tabularx}
\end{table}

Sharpe ratio is the primary ranking metric, with profit factor, maximum drawdown, and realized-return capture used as supporting diagnostics. The \emph{return capture gap} is defined as the difference between the peak unrealized return observed during a trade and the realized return at exit.

The evaluation protocol changed during analysis. The initial plan used a chronological 70/30 split in order to preserve temporal ordering between training and test trades. However, the analysis was being conducted during the opening phase of the Iran war, when the newest trades in the sample were disproportionately poor because market conditions had deteriorated sharply. Under that regime shock, the chronologically held-out segment produced extremely negative test Sharpe ratios, in some cases as low as $-5$, making the split more reflective of a single crisis window than of the broader trade distribution.

To reduce that concentration effect, the final comparative analysis randomized the trade sample before partitioning it. This choice makes the reported rankings less dominated by one adverse end-of-sample episode and more representative of the full set of observed trades. At the same time, the change weakens the interpretation of the results as a strict forward test, because randomization mixes market regimes and can leak structurally similar trades across the evaluation boundary.

Two methodological caveats matter. First, the exercise is a large parameter search on a finite historical sample, so selection bias is unavoidable \cite{bailey2014}. Second, the dataset records price paths but not all discretionary information behind original manual exits. The simulation therefore evaluates what the alternative rule set would have done on the observed path, not whether a human override was justified by external information.

\section{Results}
Table~\ref{tab:summary} summarizes the baseline, the strongest first-pass configuration, and the strongest second-pass configuration. Sharpe ratio, computed as the annualized mean return divided by the standard deviation of returns over the evaluation horizon, is the primary ranking metric. Relative to the baseline Sharpe ratio of 0.419, the best first-pass configuration reaches 0.525, a 25.2\% improvement. Adding ATR-based exits and a circuit-breaker overlay raises Sharpe to 0.653, a further 24.5\% gain over the first-pass winner and a 56.0\% gain over baseline.

\begin{table}[t]
\caption{Representative baseline and best-performing configurations.}
\label{tab:summary}
\centering
\scriptsize
\setlength{\tabcolsep}{3pt}
\renewcommand{\arraystretch}{1.08}
\begin{tabularx}{0.9\linewidth}{@{}l>{\raggedright\arraybackslash}Xccc@{}}
\toprule
Case & Configuration & Profit factor & Max DD (USD) & Sharpe \\
\midrule
Baseline & SL 0.25, TA 0.03, TD 0.02, PTP 0.05, PF 0.50, 24 h & 1.639 & 829.8 & 0.419 \\
Best pass 1 & SL 0.10, TA 0.03, TD 0.03, PTP 0.05, PF 0.75, 48 h & 1.760 & 757.2 & 0.525 \\
Best pass 2 & Base $(0.10,0.03,0.05,0.10,0.75,48)$ + ATR SL 1.0$\times$, ATR TP 2.0$\times$, CB factor 0.25 after 2 losses & 2.375 & N/A$^{\ast}$ & 0.653 \\
\bottomrule
\multicolumn{5}{@{}p{0.88\linewidth}@{}}{\scriptsize $^{\ast}$Drawdown not tracked under the ATR overlay simulator, which does not reconstruct the portfolio equity curve; only Sharpe ratio and profit factor are available from this stage.}\\
\end{tabularx}
\end{table}

The first-pass patterns are stable across the top of the ranking. Table~\ref{tab:top5} lists the five highest-Sharpe first-pass configurations. All five use a 10\% stop-loss and a 48-hour stale close. Four of the five use either a 2\% or 3\% trailing distance, and all five use a 75\% partial take-profit fraction. The configuration differences inside the top five are therefore narrow, which strengthens the claim that the broad directional pattern, rather than one fragile point estimate, is carrying the result.

\begin{table}[t]
\caption{Top five first-pass configurations by Sharpe ratio.}
\label{tab:top5}
\centering
\scriptsize
\setlength{\tabcolsep}{2.5pt}
\renewcommand{\arraystretch}{1.06}
\begin{tabular}{@{}rccccccc@{}}
\toprule
Rank & SL & TA & TD & PTP & PF & Hours & Sharpe \\
\midrule
1 & 0.10 & 0.03 & 0.03 & 0.05 & 0.75 & 48 & 0.5245 \\
2 & 0.10 & 0.03 & 0.05 & 0.10 & 0.75 & 48 & 0.5237 \\
3 & 0.10 & 0.08 & 0.02 & 0.05 & 0.75 & 48 & 0.5222 \\
4 & 0.10 & 0.05 & 0.02 & 0.05 & 0.75 & 48 & 0.5220 \\
5 & 0.10 & 0.10 & 0.02 & 0.05 & 0.75 & 48 & 0.5215 \\
\bottomrule
\end{tabular}
\end{table}

\begin{figure}[t]
\centering
\includegraphics[width=0.9\linewidth]{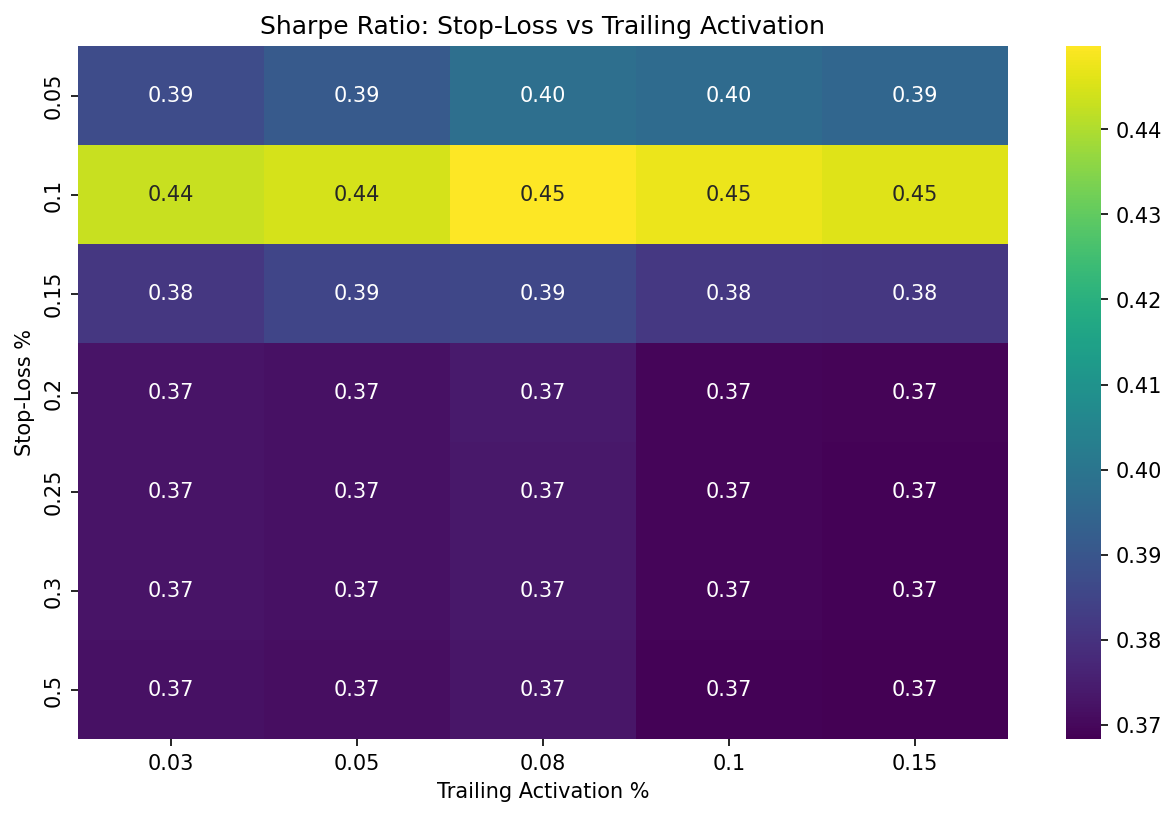}
\caption{Average Sharpe ratio by stop-loss and trailing activation.}
\label{fig:sharpe}
\end{figure}

The heatmap in Fig.~\ref{fig:sharpe} shows the same structure in aggregated form: average Sharpe is highest around a 10\% stop-loss with moderate trailing activation. This figure matters because it reduces the risk of over-reading a single top-ranked point. The higher-performing area is not isolated; it forms a coherent band in the parameter surface.

The second pass improves the frontier further. The top-ranked settings all pair the first-pass winners with the same ATR overlay, namely a 1.0$\times$ ATR stop and 2.0$\times$ ATR take-profit. The best second-pass configurations do not disable the circuit breaker; every top-five result uses the strongest reduction factor shown in the pass-2 output. The more defensible interpretation is therefore not that circuit breakers are useless, but that their value appears conditional on the specific ATR overlay and base configuration chosen.

\begin{figure}[t]
\centering
\includegraphics[width=0.9\linewidth]{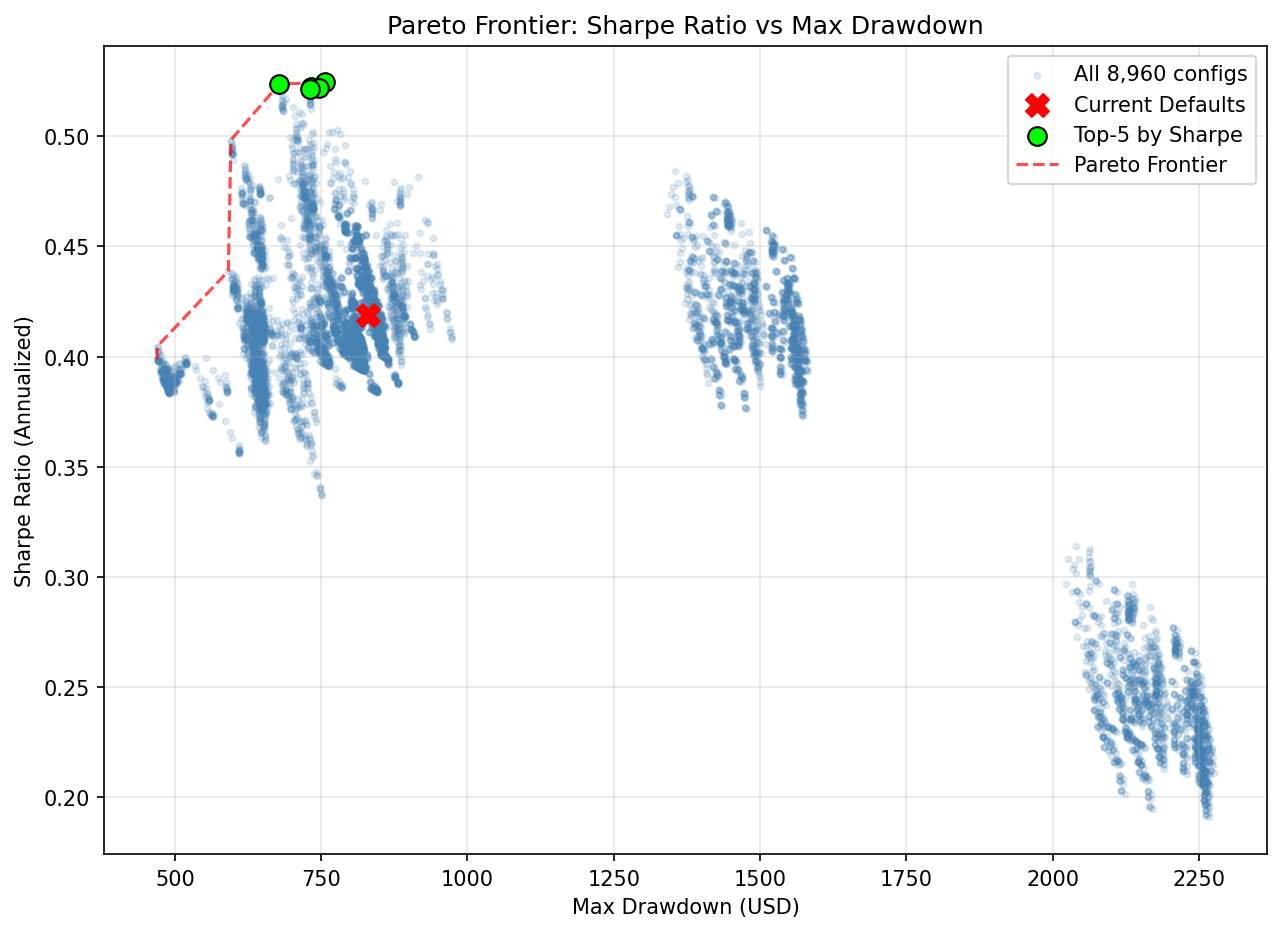}
\caption{Pareto frontier of Sharpe ratio versus maximum drawdown for the first-pass grid.}
\label{fig:pareto}
\end{figure}

The risk-return trade-off is illustrated in Fig.~\ref{fig:pareto}. The baseline lies inside the cloud of feasible configurations, whereas the leading candidates sit on or near the upper-left frontier, combining higher Sharpe with lower drawdown than many alternatives. The figure also shows that some regions of the grid produce materially worse drawdown without compensating gains in Sharpe, which supports pruning large portions of the original heuristic search space in future work.

\section{Discussion and Limitations}
The main substantive result is simple: on this dataset, the production baseline appears too permissive on losses and too slow to lock in gains. A 25\% stop-loss exposes positions to deeper adverse moves than the top-ranked alternatives, while wider trailing structures leave more profit vulnerable to reversal. Moving to a 10\% stop-loss, shortening the trailing distance, and taking partial profit earlier improves both Sharpe and profit factor.

The 48-hour stale close also deserves mention. Every first-pass top-five configuration uses the 48-hour setting, which suggests that the current 24-hour rule may truncate trades before the exit logic has time to work. That finding is empirical rather than theoretical, but it is consistent across the strongest first-pass results. Similarly, the repeated appearance of a 75\% partial take-profit fraction indicates that small early de-risking is not enough on this sample; the better-performing strategies remove more exposure when the first gain threshold is reached.

The ATR overlay provides the largest incremental improvement. A plausible mechanism is that fixed-percentage exits perform reasonably in the core search, while ATR scaling adds local volatility adaptation around those already-sensible levels. Even so, the paper should not overclaim. The original chronological split became dominated by a war-period regime break in the newest trades, driving test Sharpe to highly negative values and making the holdout look more like a concentrated crisis sample than a representative forward slice.

That observation motivated the randomized evaluation used in the final comparison, but the remedy introduces its own drawbacks. Randomization reduces the influence of a single stressed period, yet it also weakens temporal realism. In particular, it can mix highly related trades from similar market conditions across train and test partitions, understate regime-shift risk, and make the results look more stable than a true live deployment would reveal. The randomized split is therefore better viewed as a robustness-oriented sampling choice than as a clean simulation of future performance.

These limits bound the paper's contribution. The search is still effectively in-sample for practical purposes, the number of trades remains limited relative to the size of the grid, and no deflated-Sharpe correction is applied despite the relevance of selection-bias concerns \cite{bailey2014}. Recent work on reinforcement learning in finance and deep learning for trading suggests that adaptive, data-driven exit policies may offer more structural robustness than fixed-parameter grids of the kind tested here \cite{theate2021,hambly2023,zhang2020}, a direction this paper does not pursue but which the results motivate. In addition, the volatility-tier chart available in the supporting material produces implausibly negative sub-sample Sharpe values for low- and medium-volatility segments, so this revision treats volatility adaptation as a future direction rather than a validated result. Finally, the paper does not claim that the reported best configuration is the unique or globally optimal one. It is simply the strongest setting among those tested.

\section{Recommended Production Setting}
If a single deployable configuration is needed, the evidence supports the following default: 10\% stop-loss, 3\% trailing activation, 5\% trailing distance, 10\% partial take-profit threshold, 75\% partial take-profit fraction, and 48-hour stale close. Where ATR overlays are available, the strongest observed refinement is a 1.0$\times$ ATR stop and 2.0$\times$ ATR take-profit, combined with a circuit-breaker reduction factor of 0.25 after 2 consecutive losses. This recommendation is best viewed as a candidate for prospective testing, not as a final policy.

\section{Conclusion}
This paper reframes exit-rule tuning as a calibration problem rather than a heuristic choice. On a dataset of over 900 historical trades, a broad counterfactual search identifies materially stronger configurations than the production baseline, and a targeted ATR refinement stage improves the top candidate further. The qualitative lesson is stable even if the precise parameter values are not: tighter downside control, faster profit capture, and disciplined stale-close rules matter more than wide, loosely justified defaults.

The paper's limitations are substantial and should remain explicit. The findings are in-sample, the parameter search is large, and the available split diagnostics indicate instability. Even so, the study provides a cleaner and more defensible foundation for future work than the original heuristic settings. The next step should be prospective validation with walk-forward testing, explicit correction for multiple testing \cite{lopezdeprado2020}, and a volatility-segmentation analysis built on sub-samples large enough to support reliable inference.

\balance

\end{document}